\documentclass[runningheads]{llncs}
\usepackage[T1]{fontenc}
\usepackage{wrapfig,lipsum,booktabs,setspace}
\usepackage{amsmath}
\usepackage{graphicx,wrapfig,booktabs}
\PassOptionsToPackage{hyphens}{url}\usepackage{hyperref}
\usepackage[
  separate-uncertainty = true,
  multi-part-units = repeat
]{siunitx}
\usepackage{graphicx}
\begin{document}
\title{TetCNN: Convolutional Neural Networks on
Tetrahedral Meshes}
% \author{Paper ID: 195} 
\author{Mohammad Farazi\inst{1}\and
Zhangsihao Yang \inst{1}\and
Wenhui Zhu \inst{1}\and
Peijie Qiu \inst{2}\and
Yalin Wang\inst{1}}

\authorrunning{M. Farazi et al.}


\institute{School of Computing and Augmented Intelligence, Arizona State University, AZ 85281, USA \and
McKeley School of Engineering, Washington University in St. Louis, St. Louis, MO 63130, USA}

\maketitle              % typeset the header of the contribution
\begin{abstract}
Convolutional neural networks (CNN) have been broadly
studied on images, videos, graphs, and triangular meshes.  However, it has
seldom been studied on tetrahedral meshes. Given the merits of using
volumetric meshes in applications like brain image
analysis, we introduce a novel interpretable graph CNN framework for the tetrahedral mesh structure. Inspired by ChebyNet, our model exploits
the volumetric Laplace-Beltrami Operator (LBO) to define filters
over commonly used graph Laplacian which lacks the Riemannian metric information of 3D manifolds. For pooling adaptation, we introduce
new objective functions for localized minimum cuts in the Graclus algorithm based on the LBO. We employ a piece-wise constant approximation scheme that uses the clustering assignment matrix to estimate the LBO on sampled meshes after each pooling. Finally, adapting the Gradient-weighted Class Activation Mapping algorithm for tetrahedral meshes, we use the obtained heatmaps to visualize discovered regions-of-interest as biomarkers. We demonstrate the effectiveness of our model on cortical tetrahedral meshes from patients with Alzheimer’s disease, as there is scientific evidence showing the correlation of cortical thickness to neurodegenerative disease progression.
 Our results show the superiority of our LBO-based convolution layer and
adapted pooling over the conventionally used unitary cortical thickness, graph Laplacian, and point cloud representation.

\keywords{Magnetic resonance imaging \and Convolutional neural networks  \and Tetrahedral meshes \and Laplace-Beltrami operator.}
\end{abstract}
\section{Introduction}
Since the emergence of geometric deep learning research, many researchers have sought to develop learning methods on non-Euclidean domains like point clouds, surface meshes, and graphs \cite{bronstein2021geometric}. In brain magnetic resonance imaging (MRI) analysis, geometric deep learning has been widely employed for applications in brain network analysis, parcellation of brain regions, and brain cortical surface analysis\cite{fawaz2021benchmarking,zhao2021spherical,bessadok2021brain,cucurull2018convolutional}. In a benchmark study~\cite{fawaz2021benchmarking}, authors addressed the common limitations of widely used graph neural networks (GNNs) on cortical surface meshes. While the majority of these studies focus on using voxel representation and surface mesh, limitations like limited grid resolution cannot characterize complex geometrical curved surfaces precisely~\cite{Wang:NIMG17}.
% Particularly, cortical thickness changes in patients with neurodegenerative alterations to study the progression of diseases like Alzheimer's Disease (AD) \cite{ossenkoppele2019associations}.
Cortical thickness is a remarkable AD imaging biomarker; therefore, building learning-based methods will be advantageous by exploiting volumetric meshes over surface meshes since the thickness is inherently embedded in volume \cite{chandran2019supervised}. Using volumetric mesh representation also potentially helps with the over-squashing nature of Message Passing Neural Networks (MPNN) by interacting with long-range nodes through interior nodes. Thus, developing efficient volumetric deep learning methods to analyze grey matter morphometry may provide a means to analyze the totality of available brain shape information and open up new opportunities to study brain development and intervention outcomes. 

%Therefore, one may choose a 3D model that is effective in capturing the brain grey matter morphology and computationally efficient in characterizing sub-cortical areas. In case of the cortical volume, a tetrahedral mesh, generated from the inner and outer cortical surface, is a good candidate as the desired data representation. Since many neurodegenerative ilnesses are correlated with cortical thickness and changes in volume of brain cortex, a tetrahedral mesh can effectively model such changes precisely \cite{Wang:NIMG17}. 

% Although most of the works in geometric deep learning deal with graphs, networks, and point clouds, 3D mesh data are also of high interest due to their accurate, complete geometry representation capability. However, volumetric mesh representation, specifically tetrahedral mesh, is seldom studied. This type of data structure is of paramount importance in the brain imaging community, and physics-based simulation \cite{gao2020learning,Wang:NIMG17,fan2018tetrahedron,fan2021tetrahedral,zhou2020fully}. 

%Generally, in medical image analysis, volumetric details are essential in many aspects. For instance, 

While a manifold can be represented as a graph to employ graph convolutional networks, the majority of GNN models are not suitable for applying to volumetric mesh data. First, the Riemannian metric is absent in a uniform graph representation. Second and foremost, these methods mostly, are not scalable to very large sample sizes like tetrahedral meshes with millions of edges and vertices. Although there are a few methods tailored specifically to work on 3D surface meshes like MeshCNN (Convolutional Neural Networks) \cite{hanocka2019meshcnn}, these methods are particularly designed for triangular meshes and are not scalable to meshes with a high number of vertices. Consequently, to opt for a method both scalable to tetrahedral meshes and computationally inexpensive, a framework like ChebyNet~\cite{defferrard2016convolutional}, modified with volumetric LBO over graph Laplacian, is an appropriate candidate to adapt for the tetrahedral meshes. 

Although deep learning on mesh has been studied in recent years, few studies have employed explainable methods for qualitative assessment. Generally, to better explain geometric deep learning models, some methods have been proposed in recent years. Gradient-weighted Class Activation Map (Grad-CAM) on graphs \cite{pope2019explainability} is one of the first methods, using the gradient of the target task flowing back to the final convolution layer to create a localization map to visualize important nodes in the prediction task. This technique is commonplace in CNN and GNN models, however, the generalization of the concept is rarely used on surface mesh data \cite{azcona2020interpretation}. Specifically, Grad-CAM has never been investigated on volumetric meshes, and it is worth generalizing such an explainable technique for the medical image analysis community for a better interpretation of the deep learning model.

\begin{figure}[t]
\centering
\includegraphics[scale = 0.44]{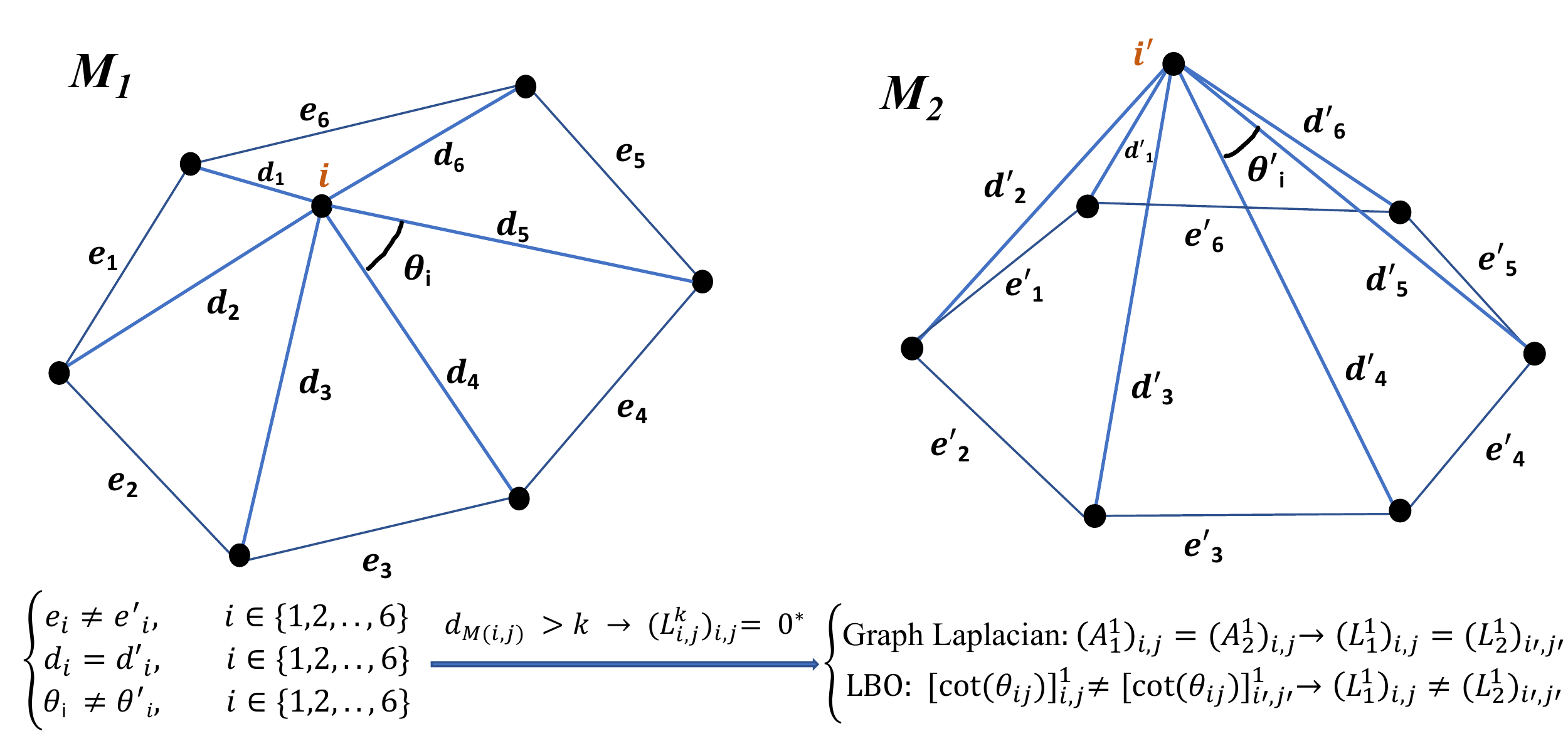}
\caption{Illustration of comparison between LBO and graph Laplacian-based spectral filters represented by $k^{th}$ order polynomials of a mesh in 1-ring neighbor of a given vertex ($i$ in $M_1$, $i'$ in $M_2$). Based on \cite{hammond2011wavelets}, the Laplacian of $k^{th}$ order polynomials are exactly $k$ localized, therefore, we use $(L^{i}_m)_{i,j}$ to show the 1-localized Laplacian and $(A^{i}_m)_{i,j}$ 1-localized adjacency matrix around vertices $i$ and $i'$ in this example. As depicted, two meshes have similar corresponding edge-length within the 1-ring of vertex $i$ and $i'$. Thus, the \textbf{1-localized} graph Laplacian of both meshes is similar while their \textbf{1-localized} LBO are different due to differences in cotangent matrix weights. The surface mesh is used for simplified intuition. $^*$ is based on Lemma $5.2$ in \cite{hammond2011wavelets}.}
\label{comp}
\end{figure}

Motivated by the prior work~\cite{defferrard2016convolutional,huang2021revisiting}, here we propose to develop Tetrahedral Mesh CNN (TetCNN) to address the issues mentioned above. Using the tetrahedral Laplace-Beltrami operator (LBO) over graph Laplacian, we use the Riemannian metric in tetrahedral meshes to capture intrinsic geometric features.  Fig.~\ref{comp} demonstrates that the LBO successfully characterizes the difference between two mesh structures while the graph Laplacian fails. %why LBO is superior in capturing geometric features . 
Additionally, we propose novel designs on the pooling layers and adopt 
the polynomial approximation~\cite{hammond2011wavelets} for computational efficiency.
%provides localized filter property. Further, adopting LBO over graph Laplacian requires our novel designs on down-sampling and pooling layers in the network. 
The main contributions of this paper, thus, are summarized as follows:
% \let\labelitemi\labelitemii
% \begin{itemize}
\textbf{(1).} TetCNN is the first of its kind and an exclusive geometric deep-learning model on tetrahedral meshes. \textbf{(2).} We use volumetric LBO to replace graph Laplacian adopted in ChebyNet~\cite{defferrard2016convolutional}. \textbf{(3).} We re-define the Graclus algorithm \cite{dhillon2007weighted} used in \cite{defferrard2016convolutional,azcona2020interpretation} by adapting a localized minimum-cut objective function using the cotangent and mass matrix. \textbf{(4).} We approximate the LBO on down-sampled mesh with the piece-wise linear approximation function. This avoids the re-computation of Laplacian in deeper layers. \textbf{(5).} We demonstrate the generalization of Grad-CAM to the tetrahedral mesh may be used for biomarker identification.
%interpretation using the Grad-CAM backbone and an up-sampling block to project the weights to the original input volume mesh. 
Our extensive experiments demonstrate the effectiveness of our proposed TCNN framework for AD research.

% We demonstrate how our LBO-based model, generally using tetrahedral mesh representation, outperforms graph Laplacian-based approaches and point cloud representations of volume data.
% \end{itemize}

\vspace{-2mm}
\section{Methods}
\vspace{-2mm}

\begin{figure}[t]
\centering
\includegraphics[width=12cm]{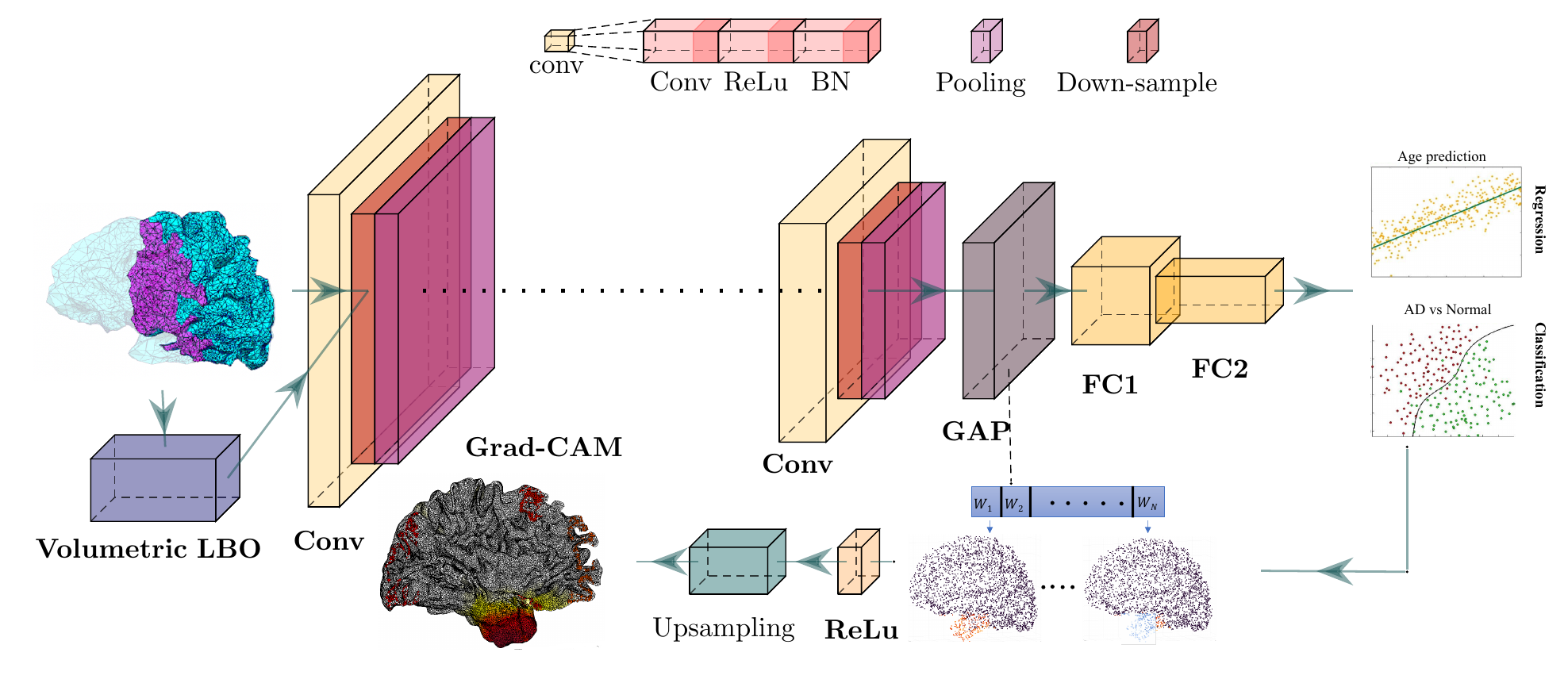}
% \vspace{-0.7cm}
\caption{TetCNN architecture for the classification task. Pre-computed LBO and $xyz$ features are fed to the network with $5$ layers. Each layer includes a down-sampling of size $1/4$ and a pooling layer afterward except for \textit{``conv5''}, which consists of a global average pooling (GAP). Fully connected (FC) layers and a Sigmoid activation function are used for the binary classification at the end. Grad-CAM is adopted to visualize important biomarkers.}\label{fig3}
\vspace{-0.2cm}
\end{figure}

%\subsection{Overview}
In our LBO-based TetCNN framework, first, we pre-compute the volumetric LBO for each tetrahedral mesh. Secondly, together with the LBO, we feed into the network a set of input features for each vertex, like the 3D coordinates of each vertex. Having built a new graph convolution layer based on the LBO, we need to down-sample the mesh with an efficient down-sampling and pooling layer to learn hierarchical feature representation for the large-sized input data. In Fig.~\ref{fig3}, we illustrate the pipeline for the binary classification task by defining specific components of our deep learning model.

% In this section, we first introduce the lumped discrete LBO to replace the graph Laplacian used in \cite{defferrard2016convolutional,ranjan2018generating,azcona2020interpretation}. Then, we explain how we build localized filters with polynomial parametrization of the tetrahedral-based LBO. Afterwards, we re-evaluate the localized min-cut objective function in \cite{dhillon2007weighted} by introducing a spectral-aware objective function using the tetrahedral LBO. Finally, we propose a simple LBO update rule for down-sampled meshes after each pooling layer.

% In 3D structured data representation like triangular mesh and tetrahedral mesh, unlike graphs, we have metric and rich information regarding the geometry and topology of the manifold. Therefore, graph neural network-based frameworks requiring graph Laplacian computation may not be an appropriate choice on a discrete manifold representation. 

% It is noteworthy to mention that non-parametric filters suffer from non-locality in space and high learning complexity of $\mathcal{O}(n^3)$. However, the adoption of polynomial approximation scheme can alleviate this computational bottleneck  \cite{defferrard2016convolutional}. 

\vspace{-4mm}
\subsection{Tetrahedral Laplace Beltrami Operator (LBO)}
Let $T$ represent the tetrahedral mesh with a set of vertices $\{v_i \}^{n}_{i=1}$ where $n$ denotes the total number of vertices, and $\Delta_{tet}$ be the volumetric LBO on $T$, which is a linear differential operator. For a Riemannian manifold, given $f \in C^2$, a real-valued function, the eigen-system of Laplacian is  $\Delta_{tet}f = -\lambda f$. The solution to this eigen-system problem can be approximated by a piece-wise linear function $f$ over the tetrahedral mesh $T$ \cite{Wang:NIMG17}. As proposed in \cite{Wang:NIMG17}, the lumped discrete LBO on $T$ is defined as follows:
\begin{equation}
    \Delta f(v_i) = \frac{1}{d_i} \sum_{j\in N(i)} k_{i,j} (f(v_i)-f(v_j))
\end{equation}
% \begin{equation}
%   S=\begin{cases}
%     \frac{1}{12}\sum_{v_j\subseteq N(v_i)}\sum_{T_l\subseteq N(v_i,v_j)}L^{(i,j)}cot\theta_l^{i,j}  , & \text{if $v_j\subseteq N(v_i)$}.\\
%     0, & \text{otherwise}.
%   \end{cases}      
% \end{equation}
where $N(i)$ includes the adjacent vertices of vertex $v_i$, $d_i$ is total tetrahedral volume of all adjacent tetrahedra to vertex $v_i$, and $k_{i,j}$ is the string constant. Now, we define the stiffness matrix as $A = W - K$ in which $W = diag(w_1,w_2,...,w_n)$ is the diagonal matrix comprised of weights $w_i =\sum_{j\in N(i)} k_{i,j} $. For $A_{ij}$ we have:

% \begin{wrapfigure}{r}{5.5cm}
% \caption{}\label{wrap-fig:1}
% \includegraphics[width=5.5cm]{tet (3).png}
% \end{wrapfigure}

% \begin{figure}[h!]

% \includegraphics[width=5cm]{tet (3).png}
% \caption{Example of a tetrahedron with dihedral angle $\theta$}
% \end{figure}
\begin{equation}
  A_{i,j}=\begin{cases}
    k_{i,j} = \frac{1}{12}\sum_{m=1}^{k} l^{(i,j)}_m cot(\theta^{(i,j)}_m)     , & \text{if $(i,j) \in E$}.\\
       0   , & \text{if $(i,j) \notin E$}.       \\
     -\sum_{q\subseteq N(i)} k_{i,q}= -\sum_{q\subseteq N(i)}\frac{1}{12}\sum_{m=1}^{k} l^{(i,q)}_m cot(\theta^{(i,q)}_m), & \text{if $i=j$},
  \end{cases}      
\end{equation}
where $l^{(i,j)}_m$ is the length of the opposite edge to $(v_i,v_j)$ in tetrahedron $m$ sharing $(v_i,v_j)$, $N(i)$ is the set of adjacent vertices to $(v_i)$, $E$ is the set of all edges in $T$, and finally $\theta^{i,j}_m$ is the diheadral angle of $(v_i,v_j)$ in tetrahedron $m$. Now, we define the lumped discrete tetrahedral LBO $L_{tet}$ given $A$ and the volume mass matrix $D$ \cite{Wang:NIMG17}:
\begin{equation}
    L_{tet} = D^{-1}A,
\label{eq3}
\end{equation}
in which $D=diag(d_1,d_2,...,d_n)$.
% \begin{equation}
%   B_{i,j}=\begin{cases}
%     \sum_{T_l\subseteq N(v_i)} \lvert\frac{V_l}{10} \rvert +  \sum_{k\subseteq N(v_i)}\sum_{T_l\subseteq N(v_i,v_k)} \lvert\frac{V_l}{20} \rvert  , & \text{if $i=j$}.\\
%      \sum_{T_l\subseteq N(v_i,v_j)} \lvert\frac{V_l}{20} \rvert              , & \text{if $v_j\subseteq N(v_i)$}.       \\
%     0, & \text{otherwise}.
%   \end{cases}      
% \end{equation}

% \begin{wrapfigure}{r}{0.48\textwidth}

% \vspace{-1.5cm}

% \includegraphics[scale = 0.18]{tet.pdf}
% \caption{A tetrahedron with dihedral angle $\theta$.}\label{wrap-fig:1}
% \end{wrapfigure}

\subsection{Spectral Filtering of Mesh Signals with Chebyshev Polynomial Approximation}
We define the input signal on the mesh as $x_{in} \in R^N$ and the output of the convolved signal with filter $g$ as $x_{out} \in R^M$. We denote the convolution operator on tetrahedral mesh $T$ with $*_T$. Following the duality property of convolution in the time domain, and having the eigenvalue and eigen-functions of tetrahedral LBO at hand, we define the convolution as:
\begin{equation}\label{eq4}
   x_{out} = g*_T x_{in} = \Phi ((\Phi^{T}g)\odot(\Phi^{T}x_{in})) =  \Phi f(\Lambda)\Phi^{T}x_{in}, 
\end{equation}
in which $\odot$ is the element-wise product, $f(\Lambda)$ is general function based on the eigen-value matrix $\Lambda$, and $\Phi$ is the eigen-vector matrix. In \cite{defferrard2016convolutional}, authors approximated the function $f$ with the linear combination of k-order power of $\Lambda$ matrix as polynomial filters:
\begin{equation}\label{eq5}
   f(\Lambda) = \sum_{m=0}^{K-1}\theta_{m}\Lambda_{tet}^{m}, 
\end{equation}
This formulation is localized in space and computationally less expensive than an arbitrary non-parametric filter $f(\Lambda)$. Per \cite{defferrard2016convolutional}, the convolution of kernel $f(.)$ centered at vertex $i$ with delta function $\delta_i$ given by $(f(L)\delta_i)_j = \sum_{k}\theta_k(L^k)_{i,j}$  gives the value at vertex $j$. Interestingly, since the $(L^k)_{i,j}$ is $K-$localized, i.e., $(L^k)_{i,j} = 0$ if $d(i,j) > K$, the locality is guaranteed with spectral filters approximated with $k-th$ polynomials of LBO \cite{defferrard2016convolutional}.
% In other perspective, the 

Now, by plugging Eq.~\ref{eq5} in Eq.~\ref{eq4}, the convolution can be expressed in terms of the Laplacian itself without any further need to calculate the eigen-functions. Chebyshev polynomials provide a boost in computational efficiency with a closed recursive formulation:
\begin{equation}\label{eq6}
   x_{out} = \sum_{m=0}^{K}\theta_{m}T_{m}(L_{tet})x_{in}, 
\end{equation}
\noindent where $\theta_{m}$ are a set of learnable model parameters denoting the coefficients of the polynomials, and $T_m \in R^{n*n}$ is the Chebyshev polynomial of order $k$.

\noindent\textbf{Recursive formulation of Chebyshev polynomials.}
The main idea of using polynomial approximation is to avoid the costly eigendecomposition and multiplication with $\Phi$. 
Therefore, we parameterize $f(\Lambda_{tet})$ with LBO, i.e., $f(L_{tet})$, 
using the recursive formulation of Chebyshev polynomials. The cost immediately reduces to $\mathcal{O}(K|\varepsilon|) \ll \mathcal{O}(n^2)$ and is desirable in graph convolution of big graphs and 3D meshes. In Eq. \ref{eq5}, the Chebyshev polynomial $T_m$ can be computed recursively using the form $T_m(x) = 2xT_{m-1}(x) - T_{m-2}$ with $T_0 = 1$ and $T_1 = x$ \cite{defferrard2016convolutional}. Here, all $T_m(k)$ create an orthonormal basis for $L^2([-1,1],\mu)$ with measure $\mu$ being $\frac{dy}{\sqrt{1-y^2}}$ in the Hilbert space of square integrable functions. Now given this recurrence, the Eq. \ref{eq6}, $T_m(L_{tet})$ is evaluated at $\Tilde{L_{tet}} = \frac{2L_{tet}}{\lambda_{max}} -I$ with the initialization of the recurrence being $\Tilde{x_0} = x_{in}$, $\Tilde{x_1} = \Tilde{L_{tet}}x_{in}$ with $\Tilde{x}$ representing $T_{m}(L_{tet})x_{in}$ in Eq. \ref{eq6}.

\subsection{Mesh Coarsening and Pooling Operation}
Although graph coarsening and mesh coarsening methods differ, using tetrahedral mesh down-sampling based on methods like Qslim \cite{garland1997surface} or learning-based methods like \cite{zhou2020fully} are both expensive and infeasible as they are template-based with registered shapes. Here we do not try to register tetrahedral meshes, and the number of vertices varies from mesh to mesh. Therefore, we propose to build a sub-sampling approach similar in \cite{defferrard2016convolutional} but using spectral-aware configuration. The method is similar to Graclus clustering by exploiting the Laplacian and matrix $D$ defined in the previous section.

\noindent\textbf{Defining the Normalized Min-Cut Based on Tetrahedral LBO.} Here, the objective function is based on normalized cut acting on vertices in a tetrahedral mesh. We need an affinity value between $(v_i,v_j)$ and $vol(.)$ to capture the volume of each node. For the volume in the normalized cut problem of a simple graph, we use the degree of the node; however, in surface and volumetric meshes, this notion refers to the area and volume of the adjacent surface and tetrahedrons of the vertex, respectively. The proposed affinity or edge distance must be correlated with the $A$ and $D$ in Eq. ~\ref{eq3}. Thus, the proposed affinity distance as a new objective function for the local normalized cut is:

\begin{equation}\label{eq7}
  d(v_i, v_j) = -A_{i,j}(\frac{1}{D_{ii}}+ \frac{1}{D_{jj}}), 
\end{equation}

Using this clustering objective function, at each step, we decimate the mesh by order of two. Consequently, $D_c(i,i)$ with $c$ denoting the coarsen graph, are updated by the sum of their weights for the new matched vertices. The algorithm repeats until all the vertices are matched. Typically, at each convolution layer, we use two or three consecutive pooling since the size of the tetrahedral mesh is very large. 

After coarsening, the challenging part is to match the new set of vertices with that of the previous ones. As proposed in  \cite{defferrard2016convolutional}, we use the same approach of exploiting a balanced binary tree and rearrangement of vertices by creating necessary fake nodes in the binary tree structure. For an exhaustive description of this approach please see Sec 2.3 in~\cite{defferrard2016convolutional}.

\noindent\textbf{Approximation of LBO on Down-sampled Mesh.}
After each pooling, we have a coarsened mesh that needs updated LBO to pass it to the new convolution layer. We adopt the piece-wise constant approximation approach \cite{liu2019spectral} where the clustering assignment matrix $G$ is used. This choice of $G$ is the most simple yet efficient one as the matrix is already computed for Graclus clustering. In some literature, they refer to $G$ as the prolongation operator. Now the updated Laplacian $\hat{L}$ can be derived using the following equation for $\hat{L}$:
\begin{equation}\label{eq8}
  \hat{L} = G^{T}LG, 
\end{equation}

\begin{figure}[t]
\centering
\includegraphics[scale =0.37]{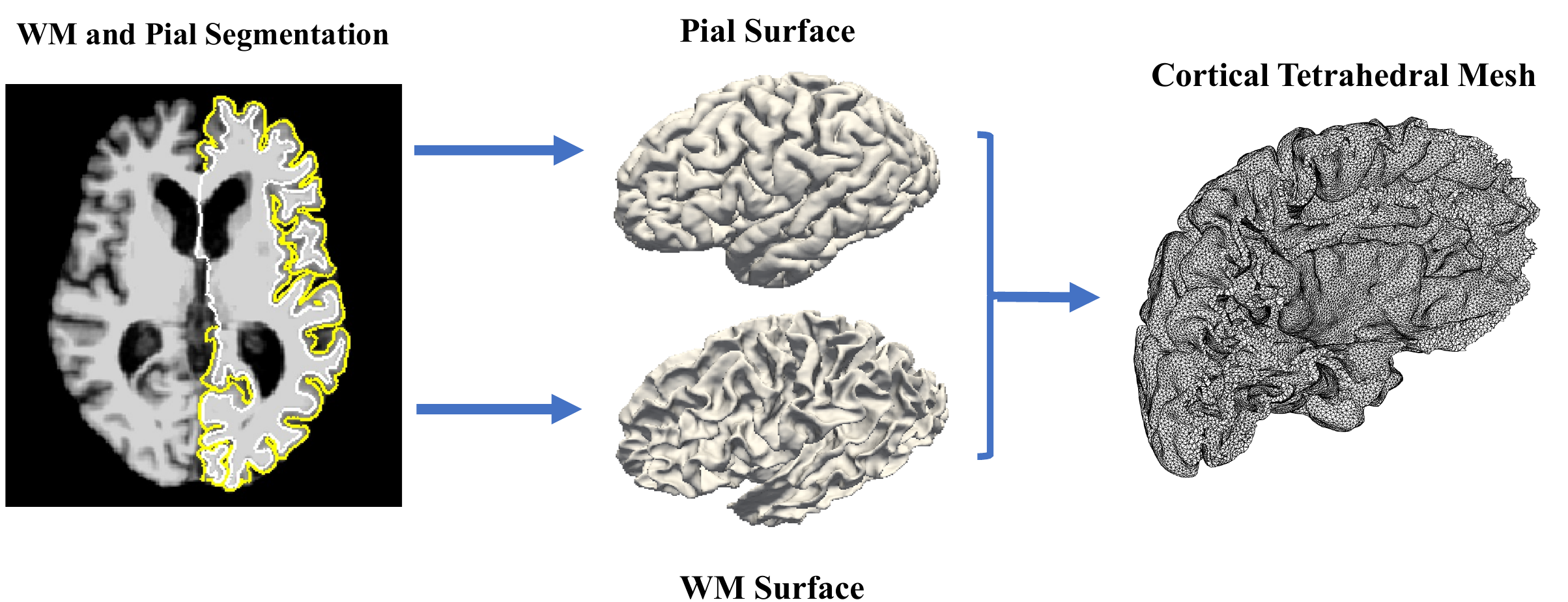}
\vspace{-0.2cm}
\caption{Procedure of creating a cortical tetrahedral mesh of a closed surface from white and pial surface pre-processed and segmented  by FreeSurfe~\cite{fischl2012freesurfer}.}\label{fig:tetmeshgeneration}
\vspace{-0.5cm}
\end{figure}

\noindent\textbf{Grad-CAM for Tetrahedral Mesh.} To utilize Grad-CAM for our framework, we need to adopt the Grad-CAM in \cite{pope2019explainability} to our tetrahedral mesh. We use the $k$-$th$ feature after the GAP layer denoted as $f_k$ which is calculated based on the last layer feature map $X^{L}_{k,n}$. Here, $n$ and $L$ refer to the $n$-$th$ node and the last layer of the network, respectively. Now, weights of Grad-CAM for class $c$ of feature $k$ in a tetrahedral mesh are calculated using 

\begin{equation}
    \alpha^{l,c}_k = \frac{1}{N} \sum_{n = 1}^{N} \frac{\partial y^{c}}{\partial X^{L}_{k,n}}
\end{equation}
To calculate the final heat map, we need to apply an activation function like ReLU and an upsampling method to project the weights to our original input mesh. As for upsampling, we use the \textit{KNN} interpolation. The final heat-map of the last layer $H$ is as follows

\begin{equation}
    H^{L,n}_c = \textrm{ReLU} (\sum_{k}\alpha^{l,c}_k X^{L}_{k,n} ) 
\end{equation}

\vspace{-6mm}
\section{Experimental Results }
\textbf{Data Processing.}
In our experiment, we study the diagnosis task for Alzheimer’s disease. Our dataset contained $116$ Alzheimer’s disease (AD) patients, and $137$ normal controls (NC) from the Alzheimer’s Disease Neuroimaging Initiative phase 2 (ADNI-2) baseline initial-visit dataset \cite{jack2008alzheimer}. All the subjects underwent the whole-brain MRI scan using a 3-Tesla MRI scanner. More details regarding the scans can be found at \url{http://adni.loni.usc.edu/wp-content/uploads/2010/05/ADNI2_GE_3T_22.0_T2.pdf}.
%The acquired MRI images are then pre-processed by FreeSurfer4 to segment pial and white matter surfaces. For mesh generation, pial and white matter surfaces are down-sampled for further tetrahedral mesh generation using the procedure stated earlier in section $2.1$.

\noindent\textbf{Cortical Tetrahedral Mesh Generation.}
We followed the procedure in \cite{fan2018tetrahedron} to create cortical tetrahedral meshes. First, pial and white surfaces were processed and created by FreeSurfer \cite{fischl2012freesurfer}. To remove self-intersections while combining pial and white surfaces, we repeatedly moved erroneous nodes and their small neighborhood along the inward normal direction by a small step size. This process continued to be done until the intersection was removed. Consequently, we used local smoothing on the modified nodes. Finally, we used TetGen \cite{si2015tetgen} to create tetrahedral meshes of the closed surfaces. Fig.~\ref{fig:tetmeshgeneration} illustrates the cortical tetrahedral mesh generation process. The number of vertices in all tetrahedral meshes was around $150k$.
To validate the robustness of our model, we used the simple \textit{xyz} coordinate as input features and normalized them using min-max normalization. We avoided using informative features as they may contribute to the final performance rather than the TetCNN itself. We pre-computed the lumped LBO for all meshes and embedded them in our customized data-loader.
\begin{table*}[t]
\centering
 \caption{Classification results between AD vs. NC under different settings and parameters (GL = graph Laplacian, (.) defines the polynomial order $k$ for LBO and GL,  E\_pool = Euclidean-based pooling). *Cortical thickness generated by FreeSurfer}\label{tab1}
\begin{tabular}{ p{3.5cm}|p{2.5cm}|p{2.5cm}|p{2.5cm} }

% \hline
% \multicolumn{5}{c}{\textbf{Classification}} \\
\hline
Method& {\textit{ACC}} & {\textit{SEN}} & {\textit{SPE}}  \\
\hline
% DGCNN & 56.2\% & - & - & - \\
% PointNet & 62.5\% & - & - & - \\
{Thickness*} & $76.2 \%$ & $77.0\% $ & $78.6\% $  \\
{LBO(1)} &  \textbf{$ \textbf{91.7}\% \pm \textbf{2.1}$} & \textbf{$89.1\% \pm 5.1$} & \textbf{$\textbf{93.3}\% \pm \textbf{3.5}$} \\
{LBO(2)} &  \textbf{$90.8\% \pm 2.0$} & \textbf{$87.5\% \pm 4.8$} & \textbf{$92.1\% \pm 3.1$}  \\ 
{GL(1)} & $87.1\% \pm 1.8$ & \textbf{$90.4\% \pm 4.7$} & $89.5\% \pm 3.1$ \\
{GL(2)}    &$85.7\% \pm 2.1$ & $\textbf{90.0}\% \pm \textbf{4.2}$& $87.5\% \pm 2.9$ \\ \hline

{LBO(1)+E\_pool}    & $84.1\% \pm 2.4$ & $83.5\% \pm 4.9$ & $87.1\% \pm 3.5$ \\
{LBO(1)+LBO\_pool} &  \textbf{$91.7\% \pm 2.1$} & \textbf{$89.1\% \pm 5.1$} & \textbf{$92.1\% \pm 3.5$} \\
\hline

\end{tabular}
\end{table*}

%%%%%%%%%%%%%%table
\begin{table*}[t]
\centering
 \caption{Classification results between AD vs. NC comparison to the baseline using different data representation. The number of different subjects is also used for fair comparison.  }\label{table_comp}
\begin{tabular}{ p{2.5cm}|p{2cm}|p{2cm}|p{2cm}|p{1.5cm} }

% \hline
% \multicolumn{5}{c}{\textbf{Classification}} \\
\hline
Study& {\textit{ACC}} & {\textit{SEN}} & {\textit{SPE}} & {\textit{Subject Split}} \\
\hline
% DGCNN & 56.2\% & - & - & - \\
% PointNet & 62.5\% & - & - & - \\

{GF-Net \cite{zhang20223d}} & $94.1\% \pm 2.8$ & $93.2\% \pm 2.4$ & $90.6\% \pm 2.6$ & (188,229) \\ 

{\textit{Qiu et al.} \cite{qiu2020development}} &  83.4\% & 76.7\% &  88.9 & (188,229) \\

{ViT3D \cite{zhang20223d}} & $85.5\% \pm 2.9$ &  $87.9\% \pm 3.6$ & $86.8\% \pm 3.7$ & (188,229) \\

{\textit{Huang et al.} \cite{huang2021revisiting}}    & $90.9\% \pm 0.6$ & $91.3\% \pm 0.1$ & $90.7\% \pm 0.5$ &(261,400) \\ 
 
{H-FCN \cite{lian2018hierarchical}} & $90.5\% $ &  $90.5\% $ & $91.3\% $ & (389,400) \\

{ResNet3D \cite{zhang20223d}} & $87.7\% \pm 3.5$ &  $90.2\% \pm 2.8$ & $89.7\% \pm 3.0$ & (188,229) \\

{DA-Net \cite{zhu2021dual}} & $92.4\% $ &  $91.0\% $ & $93.8\% $ & (389,400) \\
\hline
{\textbf{Ours}} &  $91.7\% \pm 2.1$ & $89.1\% \pm 5.1$ & $92.1\% \pm 3.5$ & (116,137)\\

\hline

\end{tabular}
\end{table*}

%%%%%%%%%%%%%%

\noindent\textbf{Classification Model Setup.}
For comparison between different manifold spectral models, we tested our model based on both tetrahedral LBO and graph Laplacian. For the sake of equal comparison among each setting, we used the same network architecture and hyper-parameters. We used $5$ TetCNN layers followed by ReLu activation function \cite{nair2010rectified} and batch-normalization \cite{ioffe2015batch}. Before the two fully connected layers, we applied a GAP to ensure the same size feature space among all mini-batches. We used $10$-fold cross-validation and picked $15\%$ of the training set for validation. We set the hyper-parameter $k$ to two different values as it is shown in Table~\ref{tab1}. The batch size for all TetCNN experiments was $8$, and the loss function used for the model was Cross-Entropy. ADAM optimizer \cite{kingma2014adam} with Learning $10^{-3}$, weight decay of $10^{-4}$, and number of epochs to $150$ were used for training the model. For the AD vs. NC classification performance evaluation, we used three measures accuracy (\textit{ACC}), sensitivity (\textit{SEN}), and specificity (\textit{SPE}). As a benchmark, we also used the FreeSurfer thickness features to train an AdaBoost classifier.     

\noindent\textbf{Point Clouds Model Setup for Classification.} Point clouds have been widely used in deep learning literature to study manifold data. In our work, we further implemented DGCNN~\cite{wang2019dynamic} and PointNet~\cite{qi2017pointnet} as our baseline models to analyze volume data. For both PointNet and DGCNN, we trained the network with batch size $1$ to feed the whole data without losing points for a fair comparison. All experiments were implemented in Python 3.7 with Pytorch Geometric 1.8 library \cite{fey2019fast} using NVIDIA GeForce Titan X GPU.

\noindent\textbf{Age Prediction Setup.}
In order to further compare the TetCNN using volumetric LBO and its Graph Laplacian counterpart, we used a regression model to see the difference in age prediction. We used the same processed data from the ADNI dataset, but we only trained the model on normal subjects. Further, we tested the trained model on both normal subjects and independent AD subjects to see the accuracy and effect of AD on age prediction. In order for an unbiased age prediction on the AD cohort, we made a test set that matched the age distribution of normal subjects. We used  $5$-$k$ fold cross-validation. As for AD subjects, we randomly chose $25$ subjects to test on the trained model and repeated the experiment $5$ times. All the parameters of the network are the same as the classification model except for the last fully connected layer the output dimension is one as we predict a number instead of discrete class labels. 

% The implementation of PointNet and DGCNN
% We implement PointNet and DGCNN as our baselines.
% For PointNet, we train the network with batch size 8 and for each point cloud we sample 48000 points from the vertices.
% For DGCNN, we need to decrease the batch size to 4 and number of points to 8000 in order to fit the network on the GPU devices.
% From our observation, both of them could not provide a comparable results comparing to our method.

\begin{table}[t]
\setlength{\tabcolsep}{24pt}
% \vspace{-1.cm}
\caption{Comparison of TetCNN with DGCNN~\cite{wang2019dynamic} and PointNet~\cite{qi2017pointnet}. }
\begin{tabular}{p{0.6cm}|p{1.5cm}|p{1.5cm}|p{1.5cm} }
\hline
 &  \textit{DGCNN}~\cite{wang2019dynamic} & \textit{PointNet}~\cite{qi2017pointnet}& \textit{TetCNN} \\

\hline
\textit{ACC}& 73.45\% & 77.35\% & \textbf{91.7\%} \\
\hline
\end{tabular}
\vspace{-0.5cm}
\label{table:ta2}
\end{table}

%%%%%fig

%%%%%%%%%%%

% \section{Results}

% \begin{table*}[t]
% \centering
%  \caption{Age prediction result......}\label{table_comp}
% \begin{tabular}{ p{2.5cm}|p{2cm}|p{2cm}}

% % \hline
% % \multicolumn{5}{c}{\textbf{Classification}} \\
% \hline
% Method& {\textit{RMSE (NC)}} & {\textit{RMSE (AD)}}  \\
% \hline
% % DGCNN & 56.2\% & - & - & - \\
% % PointNet & 62.5\% & - & - & - \\

% {LBO(1)} &  $6.2\% \pm 0.5 $& $ 7.3$ \\
% {LBO(2)} &  $6.5\% \pm 0.5$ & $7.5\% \pm 0.4 $ \\
% {GL(1)} & $7.4\% $ & $  $ \\
% {GL(2)} &  $7.8$ & $8.5$\\

% \hline

% \end{tabular}
% \end{table*}
% \vspace{-2mm}
\begin{figure*}
\centering
\includegraphics[scale=0.43,trim={1.0cm  5.2cm 1.0cm  5.1cm } ,clip]{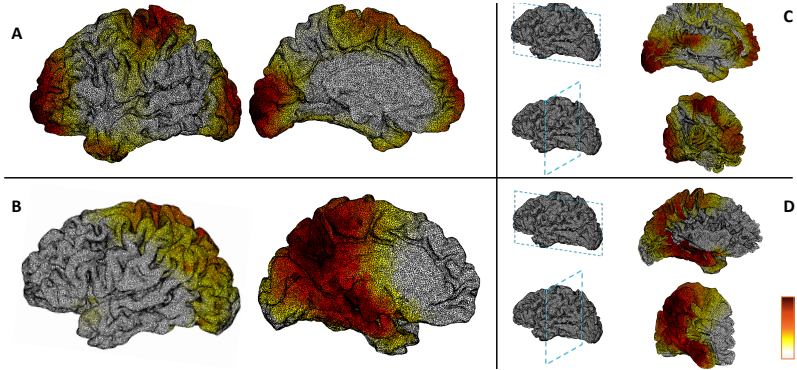}
% \vspace{-0.7cm}
\caption{Grad-CAM results for AD class showing the important regions. Comparison between LBO-based (top) and graph Laplacian (bottom) on the left hemisphere of the brain. A-B From left: Lateral-Medial view. C-D From top: Sagittal-Coronal view. Darker colors show more importance, hence greater weight.}\label{tet_grad}
\vspace{-0.2cm}
\end{figure*}
\noindent\textbf{Classification Results. } As we see in Table \ref{tab1}, TetCNN with $k=1$ outperformed any other setting, including graph Laplacian with the same parameter. 

\begin{wraptable}{r}{6cm}
 \vspace{-10pt}
\caption{Age prediction result.}\label{wrap-tab:1}

\begin{tabular}{ p{1.5cm}|p{2cm}|p{2cm}}

% \hline
% \multicolumn{5}{c}{\textbf{Classification}} \\
\hline
Method& {\textit{RMSE (NC)}} & {\textit{RMSE (AD)}}  \\
\hline
% DGCNN & 56.2\% & - & - & - \\
% PointNet & 62.5\% & - & - & - \\

{LBO(1)} &  $\textbf{6.3} \pm \textbf{0.5} yr $& $ \textbf{7.2} \pm \textbf{0.7} yr$ \\
{LBO(2)} &  $6.5 \pm 0.6 yr$ & $7.4 \pm 0.4 yr$ \\
{GL(1)} & $7.2 \pm 0.4 yr$ & $ 7.9 \pm 0.4 yr $ \\
{GL(2)} &  $7.1 \pm 0.5 yr$ & $8.1 \pm 0.5 yr$\\

\hline

\end{tabular}
 \vspace{-20pt}
\end{wraptable}
We expected the increase in $k$ would result in boosted performance, however, the results are marginally worse. We assume this behavior demonstrates the fact that $1$-ring neighbor provides sufficient information that making the receptive field larger does not contribute to more discriminative features, necessarily. Overall, TetCNN with an LBO-based setting outperformed its graph Laplacian counter-part, presumably, owing to both rich geometric features learned using LBO, as exhaustively depicted in Fig. ~\ref{comp}, and efficient spectral-based mesh down-sampling using the proposed objective function in Eq. ~\ref{eq7}. We also tested our new Graclus based on LBO and compared it to the default localized min-cut based on the Euclidean distance between two vertices. The LBO-based objective function clearly outperformed the one used in \cite{defferrard2016convolutional}, which is not suitable for mesh structure as the degree of a node is almost similar in a mesh. 

Regarding the comparison with point cloud learning frameworks, our observation in Table ~\ref{table:ta2} shows that DGCNN and PointNet could not provide comparable results with our method due to the lack of deformation sensitivity in point cloud representation. These methods produce state-of-the-art results for the classification of completely distinct objects but fail to compete with mesh structure for learning subtle deformations in volume data.

Lastly, we compared our TetCNN with other methods in the literature that are based on either brain network, surface mesh or voxel-based representations. Our results, though have a smaller dataset size for training, have comparable performance to state-of-the-art models.
%%%%%%%%%%%%%%

\noindent\textbf{Grad-CAM Results.} In Fig~\ref{tet_grad}, we illustrated the Grad-CAM results on the left grey matter tetrahedral mesh of an AD subject, trained on both LBO (A)and Graph Laplacian-based scheme (B). As illustrated, the important regions for the AD class are different in the two approaches. The identified ROIs from the LBO are more centered at the medial temporal lobe, frontal lobe, and posterior cingulate, areas that are affected by AD. But the ROIs from the graph Laplacian are more scattered, without concise ROIs. Although more validations are desired, the current results demonstrate our interpretable model may identify important AD biomarkers.  

\noindent\textbf{Age Prediction Results.} We tested our model on a regression task to compare Graph Laplacian and LBO. Furthermore, we aimed to see if the age prediction in AD patients has a larger margin of error with respect to normal subjects. Results in Table~\ref{wrap-tab:1} show the consistent outperformance of LBO-based TetCNN over its graph Laplacian counterpart. Also, it shows an erroneous prediction of AD patients with a margin of around one year which is predictable due to changes in the cortical thickness of AD patients being more severe.  

%%%%%%%%%%%
\noindent\textbf{Complexity.} Finally, in terms of computational complexity, the parameterized filter introduced in Eq. ~\ref{eq5} addresses the non-locality in space and high learning complexity of $\mathcal{O}(n)$ problem of a non-parametric filter by employing the polynomial approximation of the tetrahedral LBO.  Our novel approach reduced the time complexity to the dimension of $k$, hence $\mathcal{O}(k)$. 

 \vspace{2mm}

\section{Conclusion and Future Work }
In this study, we proposed a graph neural network based on volumetric LBO with modified pooling and down-sampling for tetrahedral meshes with different sizes. Results show the outperformance of the model to ChebyNet using graph Laplacian. Also, the adapted Grad-CAM for tetrahedral meshes showed regions affected within the surface and volume of the brain cortex in AD patients consistent with the findings in the literature. Our proposed learning framework is general and can be applied to other LBO definitions. Therefore, it may be extended to triangular mesh representation and point clouds which helped solve lots of challenging medical imaging problems including shape analysis and shape correspondence, etc. In the future, we will also study brain parcellation and segmentation tasks with our LBO-based TetCNN.

\begin{spacing}{0.90}
\bibliographystyle{splncs04}
\bibliography{egbib}
\end{spacing}
% \bibliography{egbib}
\end{document}